\documentclass[11pt,a4paper]{article}
\usepackage[hyperref]{emnlp-ijcnlp-2019}
\usepackage{times}
\usepackage{latexsym}
\usepackage{graphicx}
\usepackage{subfigure}
\usepackage{url}
\usepackage{multirow}

\aclfinalcopy 

\title{Benchmarking Zero-shot Text Classification: \\Datasets, Evaluation and Entailment Approach}

\author{Wenpeng Yin,  Jamaal Hay, Dan Roth \\
Cognitive Computation Group\\
Department of Computer and Information Science, University of Pennsylvania \\
  {\tt \small \{wenpeng,jamaalh,danroth\}@seas.upenn.edu} 
}

\date{}
\newcommand{\taskname}{\textsc{0shot-tc}}

\begin{document}
\maketitle
\begin{abstract}
Zero-shot text classification ($\taskname$) is a challenging NLU problem to which little attention has been paid by the research community. $\taskname$ aims to  
associate an appropriate label with a piece of text,  irrespective of the text domain and the aspect (e.g., topic, emotion, event,  etc.) described by the label. And there are only a few articles studying $\taskname$, all focusing only on topical categorization which, we argue, is just the tip of the iceberg in $\taskname$. In addition, the chaotic experiments in literature make no uniform comparison, which blurs the progress.

This work benchmarks the $\taskname$ problem by providing unified datasets, standardized evaluations, and state-of-the-art baselines. Our contributions include: i) The datasets we provide facilitate studying $\taskname $ relative to conceptually different and diverse aspects: the ``topic'' aspect includes ``sports'' and ``politics'' as labels; the ``emotion'' aspect includes ``joy'' and ``anger''; the ``situation'' aspect includes ``medical assistance'' and ``water shortage''. ii) We extend the existing evaluation setup (\emph{label-partially-unseen}) -- given a dataset, train on some labels, test on all labels -- to include a more challenging yet realistic evaluation \emph{label-fully-unseen} $\taskname$ \cite{DBLPChangRRS08}, aiming at classifying text snippets without seeing task specific training data at all. iii) We unify the  $\taskname$ of diverse aspects within a textual entailment formulation and study it this way.
\footnote{  \url{https://cogcomp.seas.upenn.edu/page/publication_view/883}}
\end{abstract}

\section{Introduction}
\begin{figure}[t]
\centering
\includegraphics[width=0.48\textwidth]{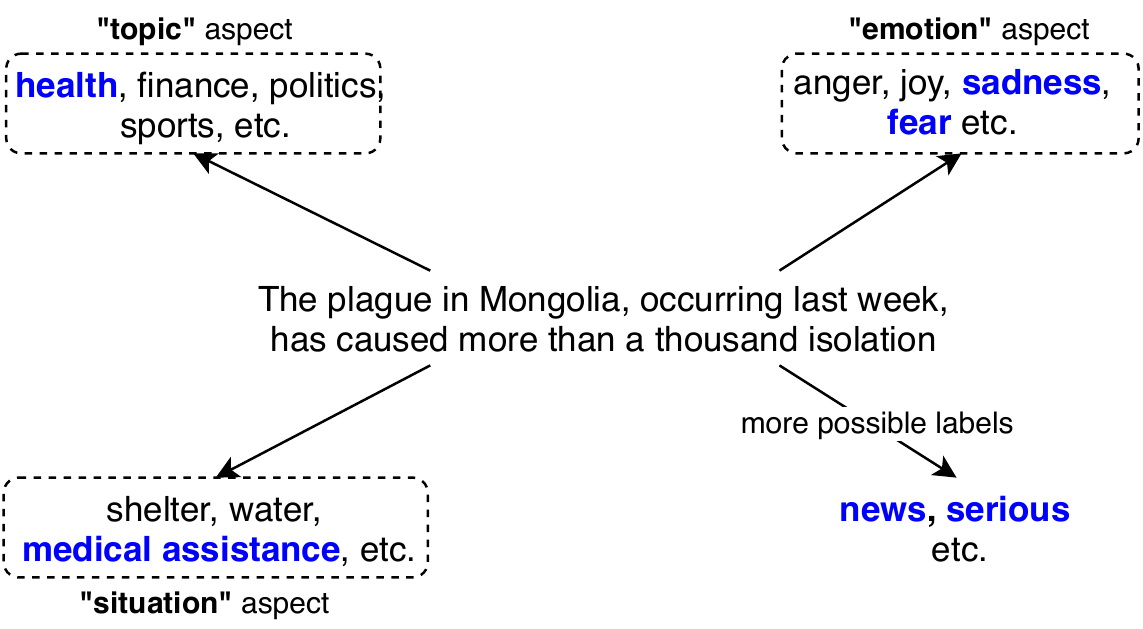}
\caption{A piece of text can be assigned labels which describe the different aspects of the text.  Positive labels are in blue.}\label{fig:zeroshot}
\end{figure}
Supervised text classification has  achieved great success in the past decades due to the availability of rich training data and deep learning techniques. However, zero-shot text classification ($\taskname$) has attracted little attention despite  its great potential in real world applications, e.g., the intent recognition of  bank consumers. $\taskname$ is challenging because we often have to deal with classes  that are compound, ultra-fine-grained, changing over time, and from different aspects such as topic, emotion, etc.

Existing $\taskname$ studies have mainly the following three problems.

\paragraph{First problem.} The $\taskname$ problem was modeled in a too restrictive vision. Firstly, most work only explored a single task, which was mainly topic categorization, e.g., \cite{DBLP171205972,DBLPYogatamaDLB17,JiangqngGuo}. We argue that this is only the tiny tip of the iceberg for $\taskname$. Secondly, there is often a  precondition that a part of classes are seen and their labeled instances are available to train a model, as we define here as \emph{Definition-Restrictive}:

\emph{Definition-Restrictive ($\taskname$)}. Given labeled  instances  belonging to a set of seen classes $S$, $\taskname$ aims at learning a classifier $f(\cdot): X \rightarrow Y$, where $Y=S\cup U$; $U$ is a set of unseen classes and belongs to the same aspect as $S$.


In this work, we formulate the $\taskname$ in a broader vision. As Figure \ref{fig:zeroshot} demonstrates, a piece of text can be assigned labels which interpret the text in different aspects, such as the ``topic'' aspect, the ``emotion'' aspect, or the ``situation'' aspect described in the text. Different aspects, therefore, differ in interpreting the text. For instance, by ``topic'', it means ``this text is about \{health, finance $\cdots$\}''; by ``emotion'', it means ``this text expresses a sense of \{joy, anger, $\cdots$\}''; by ``situation'', it means ``the people there need  \{shelter, medical assistance, $\cdots$\}''. Figure \ref{fig:zeroshot} also shows another essential property of $\taskname$ -- the applicable label space for a piece of text  has no boundary, e.g., ``this text is news'', ``the situation described in this text is serious'', etc. Therefore, we argue that we have to emphasize a more challenging scenario to satisfy the real-world problems:  seeing no labels, no label-specific training data. Here is our new $\taskname$ definition:

\emph{Definition-Wild ($\taskname$)}.  $\taskname$ aims at learning a classifier $f(\cdot): X \rightarrow Y$, where classifier $f(\cdot)$ never sees $Y$-specific labeled data in its model development.

\paragraph{Second problem.} Usually,  conventional text classification denotes labels as indices \{0,1,2, $\cdots$, $n$\} without understanding neither the aspect's specific interpretation nor the meaning of the labels. This does not apply to $\taskname$ as we can not pre-define the size of the label space anymore, and we can not presume the availability of labeled data. Humans can easily decide the truth value of any upcoming labels because humans can interpret those aspects correctly and understand the meaning of those  labels. The ultimate goal of $\taskname$ should be to develop machines to catch up with humans in this capability. To this end, making sure the system can understand the described aspect and the label meanings plays a key role.

\paragraph{Third problem.}
Prior work is mostly evaluated on different datasets and adopted different evaluation setups, which makes it hard to compare them fairly. For example, \newcite{DBLPRiosK18} work on medical data while reporting R@K as metric; \newcite{DBLPXiaZYCY18} work on SNIPS-NLU intent detection data while only unseen intents are in the label-searching space in evaluation. 

In this work, we benchmark the datasets and evaluation setups of $\taskname$. Furthermore, we propose a textual entailment approach to handle the $\taskname$ problem of diverse aspects in a unified paradigm. To be specific, we contribute in the following three aspects:
\paragraph{Dataset.} We provide  datasets for studying three aspects of $\taskname$: topic categorization, emotion detection, and situation frame detection -- an event level recognition problem. For each dataset, we have standard split for \emph{train}, \emph{dev}, and \emph{test}, and standard separation of seen and unseen classes. 

\paragraph{Evaluation.} Our standardized evaluations correspond to the \emph{Definition-Restrictive} and \emph{Definition-Wild}.  i) \emph{Label-partially-unseen evaluation}. This corresponds to the  commonly studied $\taskname$ defined in \emph{Definition-Restrictive}: for the set of labels of a specific aspect, given training data for a part of labels, predicting in the full label set. This is the most basic setup in $\taskname$. It checks whether the system can generalize to some labels in the same aspect. To satisfy  \emph{Definition-Wild}, we define a new evaluation: ii) \emph{Label-fully-unseen evaluation}. In this setup, we assume the system is unaware of the upcoming aspects and can not access any labeled data for task-specific training. 

\paragraph{Entailment approach.} Our \emph{Definition-Wild} challenges the system design -- how to develop a $\taskname$ system, without accessing any task-specific labeled data,  to deal with labels from diverse aspects? In this work, we propose to treat $\taskname$ as a textual entailment problem. This is to imitate how humans decide the truth value of labels from any aspects. Usually, humans understand the problem described by the aspect and the meaning of the label candidates. Then humans mentally construct a hypothesis by filling a label candidate, e.g., ``sports'', into the aspect-defined problem ``the text is about $\underline{?}$'', and ask ourselves if this hypothesis is true, given the text. We treat $\taskname$ as a textual entailment problem so that our model can gain knowledge from entailment datasets, and we show that it applies to both \emph{Definition-Restrictive} and \emph{Definition-Wild}.

Overall, this work aims at benchmarking the research of $\taskname$ by providing standardized datasets, evaluations, and a state-of-the-art entailment system. All datasets and codes are released.

\section{Related Work}
$\textsc{Zero-stc}$ was first explored by the paradigm 
``Dataless Classification''
\cite{DBLPChangRRS08}. 
Dataless classification first maps the text and labels into a common space by Explicit Semantic Analysis (ESA) \cite{DBLPGabrilovichM07}, then picks the label with the highest matching score. Dataless classification emphasizes that the representation of labels takes the equally crucial role  as the representation learning of text. Then this idea was further developed in \cite{DBLPSongR14,DBLPChenXJC15,DBLPLiXSM16,DBLPLiZTHIS16,DBLPSongUPR16}.

With the prevalence of word embeddings, more and more work adopts pretrained word embeddings to represent the meaning of words, so as to provide the models with the knowledge of labels \cite{DBLPSappadlaNMF16,DBLPYogatamaDLB17,DBLPRiosK18,DBLPXiaZYCY18}. \newcite{DBLPYogatamaDLB17} build generative LSTM to generate text given the embedded labels. \newcite{DBLPRiosK18} use label embedding to attend the text representation in the developing of a multi-label classifier. But they report R@K, so it is unclear whether the system can really predict unseen labels. \newcite{DBLPXiaZYCY18} study the zero-shot intent detection problem. The learned representations of intents are still the sum of word embeddings. But during testing, the intent space includes only new intents; seen intents are not covered. All of these studies can only meet the definition in \emph{Definition-Restrictive}, so they do not really generalize to open aspects of $\taskname$.

\newcite{JiangqngGuo} enrich the embedding representations by incorporating class descriptions, class hierarchy, and the word-to-label paths in ConceptNet.  \newcite{DBLPMitchellSL18} assume that some natural language explanations about new labels are available. Then those explanations are parsed into formal constraints which are further combined with unlabeled data to yield new label oriented classifiers through posterior regularization. However, those explanatory statements about new labels are collected from crowd-sourcing. This limits its application in real world $\taskname$ scenarios.



There are a few works that study a specific zero-shot problem by indirect supervision from other problems. \newcite{DBLPLevySCZ17} and \newcite{obamuyide2018zero} study zero-shot relation extraction by converting it into a machine comprehension and textual entailment problem respectively. Then, a supervised system pretrained on an existing machine  comprehension dataset or textual entailment dataset is used to do inference. Our work studies the  $\taskname$ by formulating a  broader vision: datasets of multiple apsects and evaluations. 





Other zero-shot problems studied in NLP involve entity typing \cite{DBLPZhouKTR18}, sequence labeling \cite{DBLPReiS18}, etc.

\section{Benchmark the dataset}
In this work, we standardize the datasets for $\taskname$ for three aspects: topic detection, emotion detection, and situation detection. 

For each dataset, we insist on two principles: i) \textbf{Label-partially-unseen}: A part of labels are unseen. This corresponds to \emph{Definition-Restrictive}, enabling us  to check the performance of unseen labels as well as seen labels. ii) \textbf{Label-fully-unseen}: All labels are unseen. This corresponds to \emph{Definition-Wild}, enabling us to check the system performance in  test-agnostic setups.
  

\begin{table*}
\setlength{\tabcolsep}{3pt}
  \centering
  \begin{tabular}{cc||rrrrrrrrrr|r}
  && \multicolumn{10}{c|}{emotions} & \\
  && sad& joy & anger & disgust  & fear& surp.& shame& guilt& love& none & sum\\\hline\hline
  \multirow{4}{*}{\rotatebox{90}{\begin{tabular}{c}domains\end{tabular}}} & tweets &1,500 & 2,150 &1,650 &50&2,150&880&&&1,100&1,000 & 10,480\\
  &events &300 & 200 &400 &400&200& & 300 & 300 &&& 2,100\\
  &fairytales &300 &500 &250&120 &250&220 &&&&1,000 & 2,640\\
  &arti. sent. & 200& 150 &200&30&100&100 &&&&& 780\\\hline
  \multicolumn{2}{c|}{sum}& 2,300 & 3,100 & 2,500 & 600  &2,700 & 1,200 & 300 & 300 & 1,100 & 2,000 & 16,000
\end{tabular}
\caption{Emotion \emph{test}  in $\taskname$}\label{tab:testemotion}
\end{table*}

\begin{table*}
\setlength{\tabcolsep}{3pt}
  \centering
  \begin{tabular}{cc||rrrrrrrrrr|r}
  && \multicolumn{10}{c|}{emotions} & \\
  && sad& joy & anger & disgust  & fear& surp.& shame& guilt& love& none & sum\\\hline\hline
  \multirow{4}{*}{\rotatebox{90}{\begin{tabular}{c}domains\end{tabular}}} & tweets &900&1,050&400&40&1,200&370&&&400&500 & 4,860\\
  &events &150&150&150&150&150&&100 & 100 & & & 950\\
  &fairytales &150&300&150&90&150&80&&&&500 & 1,420\\
  &arti. sent. & 100&100&100&20&100&50 &&&&& 470\\\hline
  \multicolumn{2}{c|}{sum}& 1,300 & 1,600 & 800 & 300 &1,600& 500 & 100 & 100 & 400 & 1,000 & 7,700
\end{tabular}
\caption{Emotion \emph{dev}  in $\taskname$}\label{tab:devemotion}
\end{table*}

\subsection{Topic detection}
\paragraph{Yahoo.} We use the large-scale Yahoo dataset released by \newcite{DBLPZhangZL15}. Yahoo has 10 classes: \{``Society \& Culture'',
``Science \& Mathematics'',
``Health'',
``Education \& Reference'',
``Computers \& Internet'',
``Sports'',
``Business \& Finance'',
``Entertainment \& Music'',
``Family \& Relationships'',
``Politics \& Government''\}, with original split: 1.4M/60k in train/test (all labels are balanced distributed).

We reorganize the dataset by first fixing the \emph{dev} and \emph{test} sets as follows: for \emph{dev}, all 10 labels are included, with 6k labeled instances for each; For \emph{test}, all 10 labels are  included, with 10k instances for each. Then training sets are created on remaining instances as follows. 

For \emph{label-partially-unseen}, we create two versions of Yahoo \emph{train} for $\taskname$:
\begin{itemize}
    \item \emph{Train-v0}: 5 classes: \{``Society \& Culture'', ``Health'', ``Computers \& Internet'', ``Business \& Finance'', ``Family \& Relationships''\} are included; each is equipped with  130k labeled instances.

    \item \emph{Train-v1}: 5 classes: \{
``Science \& Mathematics'', ``Education \& Reference'', ``Sports'', ``Entertainment \& Music'', ``Politics \& Government''\} are included; each is equipped with  130k labeled instances.
\end{itemize}

We always create two versions of \emph{train} with non-overlapping labels so as to get rid of the model's over-fitting on one of them. 

\emph{Label-fully-unseen}  share the same \emph{test} and \emph{dev} with the \emph{label-partially-unseen}  except that it has no training set.  It is worth mentioning that our setup of \emph{label-partially-unseen} and \emph{label-fully-unseen} enables us to compare the performance mutually; it can show the system's capabilities while seeing different sizes of classes. 


    


\subsection{Emotion detection}\label{sec:emotion}
\paragraph{UnifyEmotion.} This emotion dataset was released by \newcite{DBLPBostanK18}. It was constructed by unifying the emotion labels of  multiple public emotion datasets\footnote{Please refer to \cite{DBLPBostanK18} for more details about the constituent datasets.}. This dataset consists of text from multiple domains: tweet, emotional events, fairy tale and artificial sentences, and it contains 9 emotion types \{``sadness'', ``joy'', ``anger'', ``disgust'', ``fear'', ``surprise'', ``shame'', ``guilt'', ``love''\} and ``none'' (if no emotion applies). 
We remove the multi-label instances (appro. 4k) so that the remaining instances always have a single positive label. The official evaluation metric is \emph{label-weighted} F1. 

Since the labels in this dataset has unbalanced distribution. We first directly list the fixed $\emph{test}$ and $\emph{dev}$ in Table \ref{tab:testemotion} and Table \ref{tab:devemotion}, respectively. They are shared by following \emph{label-partial-unseen} and \emph{label-fully-unseen} setups of \emph{train}.

\emph{Label-partial-unseen} has the following two versions of  \emph{train}:
\begin{itemize}
    \item \emph{Train-v0}: 5 classes: \{``sadness'', ``anger'', ``fear'', ``shame'', ``love''\} are included.

    \item \emph{Train-v1}: 4 classes: \{
``joy'', ``disgust'', ``surprise'', ``guilt''\} are included.
\end{itemize}

For \emph{label-fully-unseen}, no training set is provided.

\subsection{Situation detection}
The situation frame typing is one example of an event-type classification task. A situation frame studied here is  a \emph{need} situation such as the need for water or medical aid, or an \emph{issue} situation such as crime violence  \cite{DBLPStrasselBT17,muisetal2018low}. It was originally designed for low-resource situation detection, where annotated data is  unavailable. This is why it is particularly suitable for $\taskname$.

We use the Situation Typing dataset released by \newcite{mayhewuniversity}. It has 5,956 labeled instances. Totally 11 situation types: ``food supply'', ``infrastructure'', ``medical assistance'', ``search/rescue'', ``shelter'', ``utilities, energy, or sanitation'', ``water supply'', ``evacuation'', ``regime change'', ``terrisms'', ``crime violence'' and an extra type ``none'' -- if none of the 11 types applies. This dataset is a \emph{multi-label} classification, and \emph{label-wise weighted F1} is the official evaluation. 

The \emph{train}, \emph{test} and \emph{dev} are listed in Table \ref{tab:sitsplit}. 

\paragraph{Summary of $\taskname$ datasets.} Our three datasets covers single-label classification (i.e., ``topic'' and ``emotion'') and multi-label classification (i.e., ``situation''). In addition, a ``none'' type is adopted in ``emotion'' and ``situation'' tasks if no predefined types apply -- this makes the problem more realistic. 


\begin{table*}
\setlength{\tabcolsep}{3pt}
  \centering
  \begin{tabular}{cc||rrrrrrrr|rrr|r}
  && \multicolumn{12}{c}{situations} \\
  && search& evac & infra & utils & water & shelter& med& food&  reg. & terr. & crim. & none\\\hline\hline
    &total size &327&278&445&412&492&659&1,046&810&80&348&983&1,868 \\\hline
\multirow{4}{*}{\rotatebox{90}{\begin{tabular}{c}split\end{tabular}}}& test &190 &166&271&260&289&396&611&472&51&204&590&1,144\\
 & dev &137&112&174&152&203&263&435&338&29&144&393&724 \\
 & train-v0&327&--&445&--&492&--&1,046&--&80&--&983&--\\
 &train-v1&--&278&--&412&--&659&--&810&--&348&--&--\\
\end{tabular}
\caption{Situation \emph{train}, \emph{dev} and \emph{test} split for $\taskname$.}\label{tab:sitsplit}
\end{table*}

\section{Benchmark the evaluation}
How to evaluate a $\taskname$ system? This needs to review the original motivation of doing $\taskname$ research. As we discussed in Introduction section, ideally, we aim to build a system that works like humans -- figuring out if a piece of text can be assigned with an open-defined label, without any constrains on the domains and the  aspects described by  the labels. Therefore, we challenge the system in two setups: \emph{label-partially-unseen} and  \emph{label-fully-unseen}. 

\paragraph{Label-partially-unseen.}
This is the most common setup in existing $\taskname$ literature: for a given dataset of a specific problem such as topic categorization, emotion detection, etc, train a system on a part of the labels, then test on the whole label space. Usually all labels describe the same aspect of the text. 


\begin{table*}
\setlength{\tabcolsep}{3pt}
  \centering
  \begin{tabular}{c|c|c|c|cc}

aspect & labels & interpretation & \multicolumn{2}{c}{example hypothesis}\\\cline{4-6}
&&&word & wordnet definition\\\hline\hline
\multirow{2}{1.5cm}{topic} & \multirow{2}{1cm}{sports\\etc.} & \multirow{2}{3.5cm}{this text is about $\underline{?}$} & \multirow{2}{2cm}{``?''= sports} & \multirow{2}{6cm}{``?'' = an active diversion requiring physical exertion and competition}\\
& & & & &\\\hline
\multirow{3}{1.5cm}{emotion} & \multirow{3}{1cm}{anger\\etc.} & \multirow{3}{3.5cm}{this text expresses $\underline{?}$} & \multirow{3}{2cm}{``?''= anger} & \multirow{3}{6cm}{``?'' = a strong emotion; a feeling that is oriented toward some real or supposed grievance}\\
& & & & &\\
& & & & &\\\hline
\multirow{2}{1.5cm}{situation} & \multirow{2}{1cm}{shelter\\etc.} & \multirow{2}{3.5cm}{The people there \\need $\underline{?}$} & \multirow{2}{2cm}{``?''= shelter} & \multirow{2}{6cm}{``?'' = a structure that provides privacy and protection from danger}\\
& & & & &\\\hline

\end{tabular}
\caption{Example hypotheses we created for modeling different aspects of $\taskname$. }\label{tab:hypoexample}
\end{table*}

\paragraph{Label-fully-unseen.}
In this setup, we push ``zero-shot'' to the extreme -- no annotated  data for any labels. So, we imagine that learning a system through whatever approaches, then testing it on $\taskname$ datasets of open aspects. 

This \emph{label-fully-unseen} setup is more like the dataless learning principle \cite{DBLPChangRRS08}, in which no task-specific annotated data is provided for training a model (since usually this kind of model fails to generalize in other domains and other tasks), therefore, we are encouraged to learn models with open-data or test-agnostic data. In this way, the learned models behave more like humans.

\section{An entailment model for $\taskname$}
As one contribution of this work, we propose to deal with $\taskname$ as a textual entailment problem. It is inspired by: i) text classification is essentially a textual entailment problem. Let us think about how humans do classification: we mentally think ``whether this text is about sport?'', or ``whether this text expresses a specific feeling?'', or ``whether the people there need water supply?'' and so on. The reason that conventional text classification did not employ entailment approach is it always has pre-defined, fixed-size of classes equipped with annotated data. However, in $\taskname$, we can neither estimate how many and what classes will be handled nor have annotated data to train class-specific parameters. Textual entailment, instead, does not preordain the  boundary of the hypothesis space. ii) To pursue the ideal generalization of classifiers, we definitely need to make sure that the classifiers understand the problem encoded in the aspects and understand the meaning of labels. Conventional supervised classifiers fail in this aspect since label names are converted into indices -- this means the classifiers do not really understand the labels, let alone the problem. Therefore, exploring $\taskname$ as a textual entailment paradigm is a reasonable way to achieve generalization. 

\paragraph{Convert labels into hypotheses.} The first step of dealing with $\taskname$ as an entailment problem is to convert labels into hypotheses. To this end, we first convert each aspect into an \emph{interpretation} (we discussed before that generally one aspect defines one interpretation). E.g.,  ``topic'' aspect to interpretation ``the text is about the topic''. Table \ref{tab:hypoexample} lists some examples for the three aspects: ``topic'', ``emotion'' and ``situation''.  

In this work, we just explored two simple methods to generate the hypotheses. As Table \ref{tab:hypoexample} shows, one is to use the label name to complete the interpretation, the other is to use the label's definition in WordNet to complete the interpretation. In testing, once one of them results in an ``entailment'' decision, then we decide the corresponding label is positive. We can definitely create more natural hypotheses through crowd-sourcing, such as ``food'' into ``the people there are starving''. Here we just set the baseline examples by automatic approaches, more explorations are left as future work, and we welcome the community to contribute.

\paragraph{Convert classification data into entailment data.} For a data split (\emph{train}, \emph{dev} and \emph{test}), each input text, acting as the premise, has a positive hypothesis corresponding to the positive label, and all negative labels  in the data split provide negative hypotheses. Note that unseen labels do not provide negative hypotheses for instances in \emph{train}.

\paragraph{Entailment model learning.} In this work, we make use of the  widely-recognized state of the art entailment technique -- BERT \cite{DBLP04805}, and train it on three mainstream entailment datasets: MNLI \cite{DBLPWilliamsNB18}, GLUE RTE \cite{DBLPDaganGM05,DBLPWangSMHLB19} and FEVER\footnote{FEVER is an evidential claim verification problem: given a hypothesis, the system needs to identify evidence sentences as premise, then gives the entailment decision. We use the ground truth evidence as premises in this work.} \cite{DBLPhorneVCM18}, respectively. We convert all datasets into binary case: ``entailment'' vs. ``non-entailment'', by changing the label ``neutral'' (if exist in some datasets) into ``non-entailment''.

For our \emph{label-fully-unseen}  setup, we directly apply this pretrained entailment model on the test sets of all $\taskname$ aspects. For \emph{label-partially-unseen} setup in which we intentionally provide  annotated data, we first pretrain BERT on the MNLI/FEVER/RTE, then fine-tune  on the provided training data. 
\begin{table*}
\setlength{\tabcolsep}{3pt}
  \centering
  \begin{tabular}{ll|cc|cc|cc|cc|cc|cc}
 & &  \multicolumn{4}{c|}{topic} & \multicolumn{4}{c|}{emotion} & \multicolumn{4}{c}{situation}\\
 && \multicolumn{2}{c|}{v0} & \multicolumn{2}{c|}{v1} & \multicolumn{2}{c|}{v0} & \multicolumn{2}{c|}{v1} & \multicolumn{2}{c|}{v0} & \multicolumn{2}{c}{v1} \\
 && s & u & s & u & s & u& s & u& s & u& s & u\\\hline\hline
\multirow{3}{*}{\rotatebox{90}{\begin{tabular}{c}w/o \\ \emph{train}\end{tabular}}} &  Majority & 0.0 & 10.0 & 10.0 & 0.0 & 0.0  & 13.3  & 18.5 & 0.0 &0.0 &19.7&0.0&16.4\\
& Word2Vec &28.1&43.3&43.3&28.1&8.1&5.4&6.2&7.3&10.3&24.7&8.6&23.1\\
& ESA &27.5&30.1&30.1&27.5&6.7&9.7&5.5&9.2&22.8&28.5&22.4&27.7\\\hline
\multirow{5}{*}{\rotatebox{90}{\begin{tabular}{c}supervised\\train\end{tabular}}} &  Binary-BERT &\textbf{72.6} &44.3&\textbf{80.6}&34.9&\textbf{35.6}&17.5&\textbf{37.1}&14.2&72.4&48.4&63.8 & 42.9\\
& our entail \\
& \enspace\enspace MNLI &70.9&\textbf{52.1}&77.3&\textbf{45.3}&33.4&\textbf{26.6}&33.9&21.4&\textbf{74.8} & \textbf{53.4} & \textbf{68.4} & \textbf{47.8}\\
& \enspace\enspace FEVER &70.2&51.7&77.2&42.7&31.9&24.5&26.0&\textbf{22.5}&73.5 & 47.6 & 65.7 & 43.6\\ 
& \enspace\enspace RTE &71.5&45.3&78.6&40.6&32.0&21.8&32.7&21.1&72.8 & 52.0&65.0 & 45.2\\\hline
\end{tabular}
\caption{Label-partially-unseen evaluation. ``v0/v1'' means the results in that column are obtained by training on train-v0/v1.  ``s'': seen labels; ``u'': unseen labels. ``Topic'' uses \emph{acc.}, both ``emotion'' and ``situation'' use \emph{label-wised weighted F1}. Note that for baselines ``Majority'', ``Word2Vec'' and ``ESA'', they do not have \emph{seen} labels; we just separate their numbers into \emph{seen} and \emph{unseen} subsets of supervised approaches for clear comparison.}\label{tab:withincorpus}
\end{table*}
\paragraph{Harsh policy in testing.} Since seen labels have annotated data for training, we adopt  different policies to pick up seen and unseen labels. To be specific, we pick a seen label with a harsher rule: i) In single-label classification,  if both seen and unseen labels are predicted as positive, we pick the seen label  only if its probability of being positive is higher than that of the unseen label by a hyperparameter $\alpha$. If only seen or unseen labels are predicted as positive, we pick the one with the highest probability; ii) In multi-label classification, if both seen and unseen labels are predicted as positive, we change the seen labels into ``negative'' if their probability of being positive is higher than that of the unseen label  by less than $\alpha$. Finally, all labels labeled positive will be selected. If no positive labels, we choose ``none'' type.

$\alpha$ = 0.05 in our systems, tuned on \emph{dev}. 

\begin{table}
\setlength{\tabcolsep}{3pt}
  \centering
  \begin{tabular}{l|ccc|r}
 & topic & emotion &  situation & sum\\\hline\hline
  Majority & 10.0 & 5.9 & 11.0 & 26.9\\
 Word2Vec &35.7 & 6.9& 15.6 & 58.2\\
 ESA &28.6 &8.0 & 26.0 & 62.6\\\hline

  Wiki-based &\textbf{52.1} &21.2&27.7 & 101.0\\
 our entail. &&&\\
 \enspace\enspace\enspace MNLI &37.9&22.3&15.4 & 75.6\\
\enspace\enspace\enspace FEVER &40.1&24.7& 21.0 & 85.8\\
 \enspace\enspace\enspace RTE &43.8&12.6& 37.2 & 93.6\\
 \enspace\enspace\enspace ensemble &45.7&\textbf{25.2}&\textbf{38.0} & \textbf{108.9}\\\hline
\end{tabular}
\caption{Label-fully-unseen evaluation.}\label{tab:hybridresults}
\end{table}
\section{Experiments}




\begin{table*}
\setlength{\tabcolsep}{2.5pt}
  \centering
  \begin{tabular}{l||cccc|cccc|cccc||cccc}
  & \multicolumn{4}{c|}{topic} & \multicolumn{4}{c|}{emotion} & \multicolumn{4}{c||}{situation}& \multicolumn{4}{c}{sum}\\\hline\hline
  
  & RTE & FEV. & MN. & ens. & RTE & FEV. & MN. & ens. & RTE & FEV. & MN. & ens.& RTE & FEV. & MN. & ens.\\\hline

word &44.9&42.0&43.4&48.4&12.4&26.7&21.2&18.3&37.7 & 24.5 & 14.7 &38.3 & 95.0&93.2&79.3 &105.0\\
def &14.5&25.3&17.2&26.0&3.4&18.7&16.8&9.0&14.1 & 19.2 & 11.8 & 14.4 & 32.0 &63.2 & 45.8 & 49.4\\
comb. &43.8&40.1&37.9&45.7&12.6&24.7&22.3&25.2&37.2 & 21.0 & 15.4 & 38.0 & 93.6 & 85.8 & 81.2 & \textbf{108.9}\\\hline
\end{tabular}
\caption{Fine-grained \emph{label-fully-unseen} performances of different hypothesis generation approaches ``word'', ``def'' (definition) and ``comb'' (word\&definition) on the three tasks (``topic'', ``emotion'' and ``situation'') based on three pretrained entailment models (RTE, FEVER, MNLI) and the ensemble approach (ens.). The last column \emph{sum} contains the addition of its preceding  three blocks element-wisely.}\label{tab:finegrainedhypo}
\end{table*}
\subsection{Label-partially-unseen evaluation}
In this setup, there is annotated data for partial labels as \emph{train}. So, we report performance for unseen classes as well as seen classes. We compare our entailment approaches, trained separately on MNLI, FEVER and RTE,  with the following baselines. 
\paragraph{Baselines.} \begin{itemize}
    \item \emph{Majority}: the text picks the label of the largest size.
    \item \emph{ESA}: A dataless classifier proposed in \cite{DBLPChangRRS08}. It maps the words (in text and label names) into the title space of Wikipedia articles, then compares the text with label names. This method does not rely on \emph{train}.
    
    We implemented ESA based on 08/01/2019 Wikipedia dump\footnote{\url{https://dumps.wikimedia.org/enwiki/}}. There are about 6.1M words and 5.9M articles.  
    \item \emph{Word2Vec}\footnote{\url{https://code.google.com/archive/p/word2vec/}} \cite{DBLPMikolovSCCD13}: Both the representations of the text and the labels are the addition of word embeddings element-wisely. Then cosine similarity determines the labels. This method does not rely on \emph{train} either.
    \item \emph{Binary-BERT}: We fine-tune BERT\footnote{We always use ``bert-base-uncased'' version.} on \emph{train}, which will yield a binary classifier for entailment or not; then we test it on \emph{test} -- picking the label with the maximal probability in single-label scenarios while choosing all the labels with ``entailment'' decision in multi-label cases. 
\end{itemize}

\paragraph{Discussion.} The results of \emph{label-partially-unseen} are listed in Table \ref{tab:withincorpus}. ``ESA'' performs slightly worse than ``Word2Vec'' in topic detection, mainly because the label names, i.e., topics such as ``sports'', are closer than some keywords such as ``basketball'' in Word2Vec space. However, ``ESA'' is clearly better than ``Word2Vec'' in situation detection; this should be mainly due to the fact that the label names (e.g., ``shelter'', ``evaculation'', etc.) can hardly find close words in the text by Word2Vec embeddings. Quite the contrary, ``ESA'' is easier to make a class such as ``shelter'' closer to some keywords like ``earthquake''. Unfortunately, both Word2Vec and ESA work poorly for emotion detection problem. We suspect that emotion detection requires more entailment capability. For example, the text snippet ``when my brother was very late in arriving home from work'', its gold emotion ``fear'' requires some common-knowledge inference, rather than just word semantic matching through Word2Vec and ESA. 

The supervised method ``Binary-BERT'' is indeed strong in learning the seen-label-specific models -- this is why it predicts very well for seen classes while performing much worse for unseen classes. 

Our entailment models, especially the one pretrained on MNLI, generally get competitive performance with the ``Binary-BERT'' for \emph{seen} (slightly worse on ``topic'' and ``emotion'' while clearly better on ``situation'') and  improve the performance regarding \emph{unseen} by large margins. At this stage, fine-tuning on an MNLI-based pretrained entailment model seems more powerful. 

\subsection{Label-fully-unseen evaluation}
Regarding this \emph{label-fully-unseen} evaluation, apart from our entailment models and three unsupervised baselines ``Majority'', ``Word2Vec'' and ``ESA'', we also report the following baseline:

\textbf{Wikipedia-based}: We train a \emph{binary} classifier based on BERT on a dataset collected from Wikipedia. Wikipedia is a corpus of general purpose, without targeting any specific $\taskname$ task.  Collecting categorized articles from Wikipedia is popular way of creating training data for text categorization, such as \cite{DBLPZhouKTR18}. More specifically, we collected 100K articles along with their categories in the bottom of each article. For each article, apart from its attached positive categories, we randomly sample three negative categories. Then each article and its positive/negative categories act as training pairs for the binary classifier. 

We notice ``Wikipedia-based'' training indeed contributes a lot for the topic detection task; however, its performances on emotion and situation detection problems are far from satisfactory. We believe this is mainly because the Yahoo-based topic categorization task is much closer to the Wikipedia-based topic categorization task; emotion and situation categorizations, however, are relatively further. 

Our entailment models, pretrained on MNLI/FEVER/RTE respectively,  perform more robust on the three $\taskname$ aspects (except for the RTE on emotion). Recall that they are not trained on any text classification  data, and never know the domain and the aspects in the \emph{test}. This clearly shows the great promise of developing textual entailment models for $\taskname$. Our ensemble approach\footnote{For each input pair of the entailment model, we sum up their probabilities after softmax, then do softmax to get new probabilities.} further boosts the performances on all three tasks.

An interesting phenomenon, comparing the \emph{label-partially-unseen} results in Table \ref{tab:withincorpus} and the \emph{label-fully-unseen} results in Table \ref{tab:hybridresults}, is that the pretrained entailment models work in this order for \emph{label-fully-unseen} case: RTE $>$ FEVER $>$MNLI; on the contrary, if we fine-tune them on the label-partially-unseen case, the MNLI-based model performs best. This could be due to a possibility that, on one hand,  the constructed situation entailment dataset is closer to the RTE dataset than  to the MNLI dataset, so an RTE-based model can generalize well to situation data, but, on the other hand, it could also be more likely to over-fit  the  training set of ``situation'' during fine-tuning. A deeper exploration of this is left as future work.

\subsection{How do the generated hypotheses influence}
\begin{figure}
\centering
\includegraphics[width=0.48\textwidth]{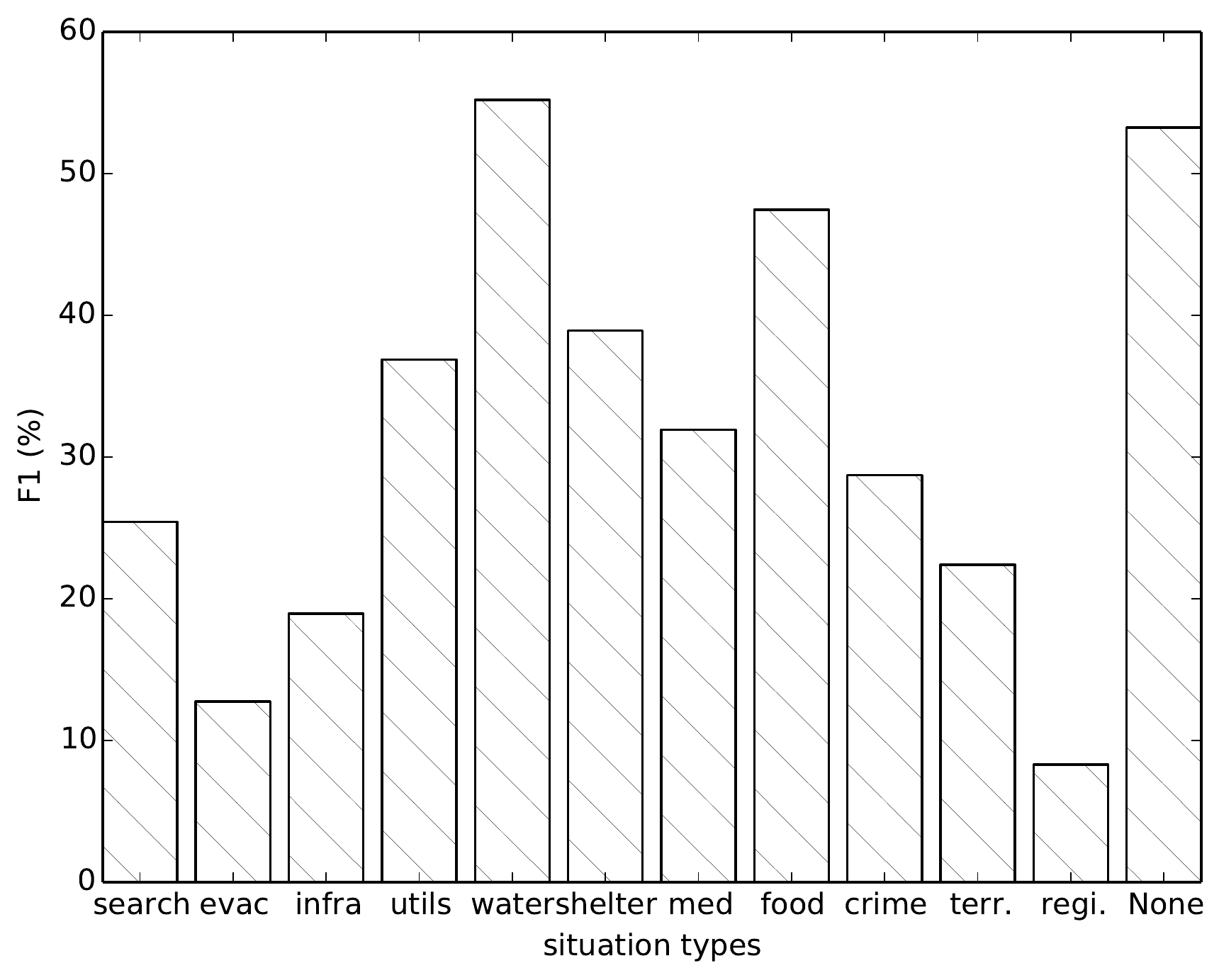}
\caption{Performance of different situation classes in \emph{label-fully-unseen, predicted by the ensemble model}.}\label{fig:hypoablation2}
\end{figure}
In Table \ref{tab:hypoexample}, we listed examples for converting class names into hypotheses. In this work, we only tried to make use of the class names and their definitions in WordNet. Table \ref{tab:finegrainedhypo} lists the fine-grained performance of three ways of generating hypotheses: ``word'', ``definition'', and ``combination'' (i.e., word\&definition). 

This table indicates that: i) Definition alone usually does not work well in any of the three tasks, no matter which pretrained entailment model is used; ii) Whether ``word'' alone or ``word\&definition'' works better depends on the specific task and the pretrained entailment model. For example, the pretrained MNLI model prefers ``word\&definition'' in both ``emotion'' and ``situation'' detection tasks. However, the other two entailment models (RTE and FEVER) mostly prefer ``word''. iii) Since it is unrealistic to adopt only one entailment model, such as from \{RTE, FEVER, MNLI\}, for any open $\taskname$ problem, an ensemble system should be preferred. However, the concrete implementation of the ensemble system also influences the strengths of different hypothesis generation approaches. In this work, our ensemble method reaches the top performance when combining the ``word'' and ``definition''. More   ensemble systems and hypothesis generation paradigms need to be studied in the future.

To better understand the impact of generated hypotheses, we dive into the performance of each labels, taking ``situation detection'' as an example.  Figure \ref{fig:hypoablation2} illustrates the separate F1 scores for each situation class, predicted by the ensemble model for \emph{label-fully-unseen} setup. This enables us to check in detail how easily the constructed hypotheses can be understood by the entailment model. Unfortunately, some classes are still challenging, such as ``evacuation'', ``infrastructure'', and ``regime change''. This should be attributed to their over-abstract meaning. Some classes were well recognized, such as ``water'', ``shelter'', and ``food''. One reason is that these labels mostly are common words -- systems can more easily match them to the text; the other reason is that they are situation classes with higher frequencies (refer to Table \ref{tab:sitsplit}) -- this is reasonable based on our common knowledge about disasters.

\section{Summary}
In this work, we analyzed the problems of existing research on zero-shot text classification ($\taskname$): restrictive problem definition, the weakness in understanding the problem and the labels' meaning, and the chaos of datasets and evaluation setups. Therefore, we are benchmarking $\taskname$ by standardizing the datasets and evaluations. More importantly, to tackle the broader-defined $\taskname$, we proposed a textual entailment framework which can work with or without the annotated data of \emph{seen} labels. 

\section*{Acknowledgments}

The authors would like to thank Jennifer Sheffield and the anonymous reviewers for insightful comments and suggestions.  This work was supported by Contracts HR0011-15-C-0113 and HR0011-18-2-0052 with the US Defense Advanced Research Projects Agency (DARPA). Approved for
Public Release, Distribution Unlimited. The views
expressed are those of the authors and do not reflect
the official policy or position of the Department of
Defense or the U.S. Government. 

\bibliography{emnlp-ijcnlp-2019}
\bibliographystyle{acl_natbib}

\end{document}